\documentclass[11pt, journal, onecolumn]{IEEEtran}
\usepackage{graphicx}
\usepackage{amsmath,amssymb} % define this before the line numbering.
\usepackage{color}
\usepackage[ left=18mm, right=18mm, top=22mm, bottom=22mm]{geometry}
\usepackage{arydshln}

\newcommand{\cN}{{\cal N}}

\newcommand{\cR}{{\cal R}}

\newcommand{\cX}{{\cal X}}
\newcommand{\cY}{{\cal Y}}

\begin{document}
% \linenumbers
%\pagestyle{headings}
%\mainmatter
%\def\ECCV18SubNumber{***}  % Insert your submission number here

\title{Training Recurrent Neural Networks against Noisy Computations during Inference} % Replace with your title

%\titlerunning{}

%\authorrunning{}

\author{Minghai Qin and Dejan Vucinic\\ Western Digital Research, \{Minghai.Qin, Dejan.Vucinic\}@wdc.com\vspace{-2em}}

\maketitle

\begin{abstract}
We explore the robustness of recurrent neural networks when the computations within the network are noisy.
One of the motivations for looking into this problem is to reduce the high power cost of conventional computing of
neural network operations through the use of analog neuromorphic circuits.
Traditional GPU/CPU-centered deep learning architectures exhibit bottlenecks in power-restricted applications,
such as speech recognition in embedded systems.
The use of specialized neuromorphic circuits, where analog signals passed through memory-cell arrays are sensed to
accomplish matrix-vector multiplications, promises large power savings and speed gains but brings with it the
problems of limited precision of computations and unavoidable analog noise.

In this paper we propose a method, called {\em Deep Noise Injection training}, to train RNNs to obtain a set of
weights/biases that is much more robust against noisy computation during inference.
We explore several RNN architectures, such as vanilla RNN and long-short-term memories (LSTM),
and show that after convergence of Deep Noise Injection training the set of trained weights/biases has
more consistent performance over a wide range of noise powers entering the network during inference.
Surprisingly, we find that Deep Noise Injection training improves overall performance of some networks even
for numerically accurate inference.

\end{abstract}

\section{Introduction}

Neural networks (NNs)~\cite{Schmidhuber15,HTF01}  are one of the most widely-used machine learning
techniques due to their good performance in practice at a variety of tasks. Some variants of neural
networks were shown to be more suitable for different learning applications.
For example, recurrent neural networks
(RNN)~\cite{GLF09,HAF14} perform better at sequence prediction, e.g. speech or text recognition.
RNNs have memory units such as long-short-term-memory (LSTM)~\cite{GSKSS16} that can be trained
without vanishing/exploding gradient problems.

A neural network is defined by the connections between the neurons, each of which is associated
with a trainable parameter called a {\em weight}. There is another parameter associated with each
neuron called a {\em bias}. Since a bias can be viewed as a weight from a neuron with constant input,
we will indiscriminately call it a weight as well.
Any given set of all trainable weights suited to performing a particular task is usually acquired by
back-propagation algorithm~\cite{RB93,Nielsen89}.

In order to fit highly non-linear functions and thus to achieve a high rate of correctness in practice,
RNNs usually contain many cells, and all cells share a large number of weights. Each RNN cell receives
signals from the input source (image, audio, text, etc.) and the output of a previous cell.
It processes the combined signal and generates the output signal to the next RNN cell.
The most power- and time-consuming computation in this process is the matrix-vector
multiplication between weights and signals. 
In power-restricted applications such as image/audio/language inference engines in embedded systems,
the practical size of a RNN is limited by the power consumption of the computations on the CPU/GPU.
One attractive method of lowering this power consumption is by use of neuromorphic computing~\cite{ZADFBG10},
where analog signals passed through memory-cell arrays are sensed to accomplish matrix-vector multiplications.
Here weights are programmed as conductances (reciprocal of resistivity) of memory cells in a 2D array.
According to Ohm's law, if input signals are presented as voltages to this layer of memory cells, 
the output current is the matrix-vector multiplication of the weights and the input signal.
The massive parallelism of this operation also promises significant savings in latency over sequential digital computation.

One of the problems with neuromorphic computing is the limited precision of the computations,
since the programming of memory cells, as well as the measurement of the analog outputs, inevitably suffers
from analog noise. Some researchers have proposed to correct the errors after each matrix-vector
multiplication~\cite{FWI18,JRRR17}. This post-error-correction technique induces extra overhead
in latency, space, and power as well. We will show that it can be foregone through the use of a
specific training technique.

The robustness of deep neural networks against random and adversarial noise in input
images/speech has been studied in~\cite{CW17,ZSLG16,YLZetal2015,Andrzej14}. In this paper, all input images are assumed noise-free. To our knowledge,
there has been no other work exploring of robustness of RNNs against noisy analog and digital
computations within the network itself. 

In this paper, we note that conventionally trained RNNs are quite sensitive to noisy computations,
such that the accuracy will drop dramatically as the power of the computation noise increases.
We propose a method to train RNNs to obtain a set of weights that exhibits a more consistent
performance over a large range of noise powers injected during inference.
Our method is based on injecting noise after each matrix-vector multiplication during training.

Note that deep noise layers were previously used in convolutional neural networks (CNNs) as a
regularization tool during training, but they are not used in RNNs and are turned off during
inference since conventional deep learning on CNNs is in digital domain so that computations are
considered noiseless.  One of our contributions is to apply the noise injection method during
both training and inference of RNNs to realize that the noisy computation problem in
neuromorphic computing can be largely mitigated by this method.

In some of our experiments, the injected noise power is as large as the signal power of RNN layers in
the network. The improvement of robustness comes from a trade-off among validation accuracy for
different noise powers. In other words, by introducing noise after each matrix-vector
multiplication during training (called {\em Deep Noise Injection training}), the validation accuracy
with noisy computations is largely improved since the trained weights are adjusted to the noise,
even for the cases where the distribution of training and validation noise are different.
We test the prediction accuracy of different RNN architectures, including vanilla RNNs and LSTMs,
with noisy computations of different noise powers.  For example, for a LSTM model used in the
MNIST handwritten digit classification task, the proposed method achieves a prediction accuracy
over $98\% $ for all additive Gaussian distributed noise of powers between $0.0$ (no noise) and $1.0$,
while it decreases from around $98.7\%$ to $12.5\%$ for a conventionally trained RNN without Deep Noise injection.
Note that our ultimate goal is neither the noise-free accuracy nor the accuracy at a particular noise level; it is
desirable to obtain a set of weights that performs well over a large range of noise levels so it can
be used for neuromophic computing with high device-to-device differences and uneven sensor precisions
resulting from unavoidable manufacturing variability.

%The rest of the paper is organized as follows. %Section~\ref{sec:prelim} provides basic knowledge of neural networks and some variants. Section~\ref{sec:cnn} explores the robustness of CNNs against noisy computations and we show that Deep Noise Injection training can improve the validation accuracy with the presence of noise. Section~\ref{sec:RNNs} explores different recurrent neural networks such as LSTMs and vanilla RNNs to show the improvement of prediction accuracy of Deep Noise Injection training.

\section{Preliminaries}\label{sec:prelim}
%\subsection{Neural networks and notations}
A neural network contains input neurons, hidden neurons, and output neurons. It can be viewed as a function $f:\cX \rightarrow \cY$ where the input $x\in\cX\subseteq\cR^n$ is an $n$-dimensional vector and the output $y\in\cY\subseteq\cR^m$ is an $m$-dimensional vector. In this paper, we focus on classification problems where the output $y=(y_1,\ldots,y_m)$ is usually normalized such that $\sum_{i=1}^m y_i = 1$ and $y_i$ can be viewed as the probability for some input $x$ to be categorized as the $i$-th class. The normalization is often done by the softmax function that maps an arbitrary $m$-dimensional vector $\hat y$ into normalized $y$, denoted by $y=softmax(\hat y)$, as $y_i = \frac{\exp(\hat y_i)}{\sum_{i=1}^m \exp(\hat y_i)}, i = 1,\ldots,m$. For top-$k$ decision problems, we return the top $k$ categories with the largest output $y_i$. In particular for hard decision problems where $k=1$, the classification result is then $\arg\max_i y_i, i=1,\ldots,m$.

A feedforward neural network $f$ that contains $n$ layers (excluding the softmax output layer) can be expressed as a concatenation of $n$ functions $f_i, i = 1,2,\ldots,n$ such that $f = f_n(f_{n-1}(\cdots f_1(x)\cdots))$. The $i$th layer $f_i: \cX_i\rightarrow \cY_i$ satisfies $\cY_i\subseteq \cX_{i+1}$, $\cX_1 = \cX$. The output of last layer $\cY_n$ is then fed into the softmax function. The function $f_i$ is usually defined as
\begin{align}\label{eq:wx+b}
f_i(x) = g(W \cdot x + b), 
\end{align}
where $W$ is the weights matrix, $b$ is the bias vector, and $g(\cdot)$ is an element-wise activation function that is usually nonlinear, e.g., tanh, sigmoid,  rectified linear unit (ReLU)~\cite{HSMDS00} or leaky ReLU~\cite{MHN13}. Both $W$ and $b$ are trainable parameters.

In this paper, we assume that there exists an additive Gaussian noise $z\sim\cN(0,\sigma^2)$ in the forward pass after each matrix-vector multiplication. Eq.~$(\ref{eq:wx+b})$ then becomes
\begin{align}\label{eq:wx+b+z}
f_i(x) = g(W \cdot x + b + z), 
\end{align}

\begin{figure}[h]
	\centering
	\includegraphics[width=0.7\linewidth]{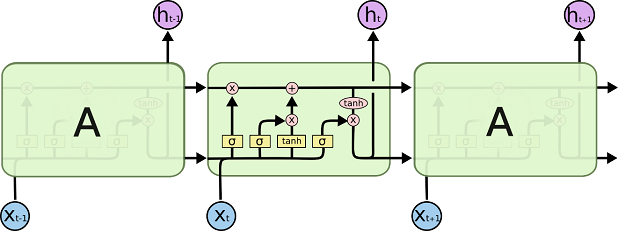}
	\caption{A LSTM cell}
	\label{fig:lstm_cell}
\end{figure}

A recurrent neural network (RNN) is a special class of neural network that has directed cycles, which enable it to create internal states and exhibit temporal behaviors. A RNN can be unfolded in time to form a feedforward neural network for training and inference. One of the most widely used neurons to store the states is LSTM (see Fig.~\ref{fig:lstm_cell}), consisting of forget-gate, update-gate, and output-gate. Gated recurrent unit (GRU) is another similar neuron style to LSTM. Back-propagation algorithms can be applied from the last output neurons backwards to train all weights in the RNN. To avoid confusion, we will call LSTM-based recurrent neural networks ``LSTMs'' for short and call the recurrent neural networks without gates ``vanilla RNNs''.
%\begin{figure}[htbp]
%	\centering
%	\includegraphics[width=0.9\linewidth]{rnn}
%	\caption{Unfolding a RNN}
%	\label{fig:rnn}
%\end{figure}

\section{Robustness of RNNs against Noisy Computations}\label{sec:RNNs}
RNNs/LSTMs are typically used for sequence prediction and classification, such as language and speech, but they are also capable of recognizing images if we sequentially feed each row of the image to one RNN/LSTM cell.  
As a result, the number of RNN/LSTM cells (called steps) equals the number of rows in the image and the input dimension of one RNN/LSTM cell equals the number of columns in the image. 
Another way to use RNNs/LSTMs for MNIST dataset is to convert the images into stroke sequences consisting of two-dimensional coordinates.
Since LSTMs have been shown to provide better performance than vanilla RNNs, we will explore the robustness of LSTMs first and then show that Deep Noise Injection training also improves the performance of vanilla RNNs as well.

\subsection{A 28-step LSTM architecture with Input Dim = 28 for MNIST}
Table~\ref{tab:lstm_mnist_input28} summarizes the LSTM architecture in our experiments. The input to each LSTM cell is a row in the image. Gaussian noise is added to five matrix-vector multiplications: four in the LSTM cell (See 4 boxes with ``$\sigma$'' and ``tanh'' in Fig.~\ref{fig:lstm_cell}) and one in the fully connected layer between the 128-dimensional LSTM output state and the 10 neurons for classification. The internal states (upper horizontal signal flow in Fig.~\ref{fig:lstm_cell}) only involve element-wise multiplication so they are not affected by the noise.

\begin{table}[htbp]
	\center
	\caption{A 28-step LSTM architecture for MNIST}
	\label{tab:lstm_mnist_input28}
	\begin{tabular}{|c|c|}
		\hline No. LSTM cells & 28 \\
		\hline Input shape & (28,1) \\
		\hline LSTM internal state size & 128\\
		\hline LSTM output state size & 128 \\
		\hline Total No. trainable parameters & $81674$\\
		\hline
	\end{tabular} 
\end{table}

\begin{figure}[h]
	\hspace{-1.4em}
	\includegraphics[width=1.15\linewidth]{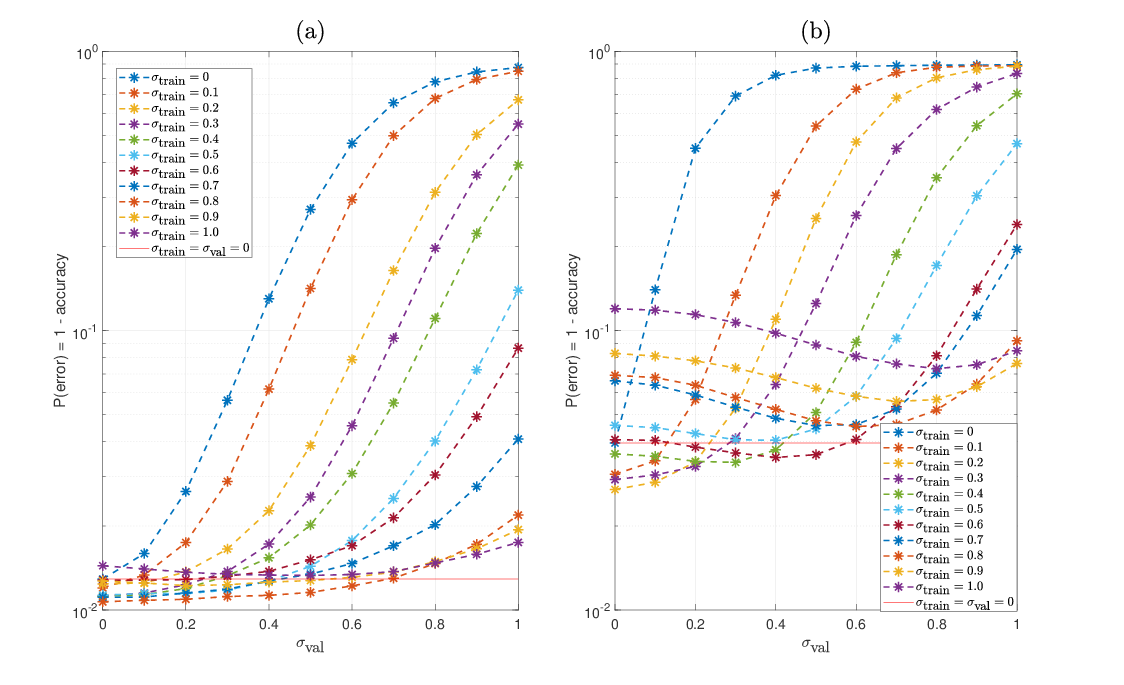}
	\caption{Prediction accuracy and error rate of LSTMs with Deep Noise Injection training of $\sigma_{\textrm{train}}=0$ to $1.0$ at a step size of $0.1$. Validated on $\sigma_{\textrm{val}}=0$ to $1.0$ at a step size of $0.1$. $(a)$ shows the accuracy of architecture in Table~\ref{tab:lstm_mnist_input28} where the input is 28-dimensional rows of images, $(b)$ shows the accuracy of architectures in Table~\ref{tab:lstm_mnist_input2} where the input is 2-dimensional coordinates of pen stroke sequences.}
	\label{fig:lstm_mnist_image_and_stroke_noisy_training}
\end{figure}

Fig.~\ref{fig:lstm_mnist_image_and_stroke_noisy_training}$(a)$ shows the prediction accuracy of $(\sigma_{\textrm{train}},\sigma_{\textrm{val}})$ pairs with $\sigma_{\textrm{train}}=0.0$ to $1.0$ and $\sigma_{\textrm{val}}=0.0$ to $1.0$, both with a step size of $0.1$. Each point is obtained as the average of $40$ independent tests. The top curve corresponds to conventional training without noise injection.
It shows the least robustness against noisy computations: validation accuracy drops from $98.7\%$ to $12.5\%$ as the noise power increases from $0.0$ to $1.0$. Increasing $\sigma_{\textrm{train}}$ generally results in higher robustness; in particular, when $\sigma_{\textrm{train}}=1.0$ validation accuracy at all noise levels during inference is greater than $98\%$.

The LSTM architecture provides an interesting observation that there exist $(\sigma_{\textrm{train}},\sigma_{\textrm{val}})$ pairs with prediction accuracy better than conventional training and validation ($\sigma_{\textrm{train}}=\sigma_{\textrm{val}}=0$). This can be seen by observing the points below the horizontal red line. This phenomenon is to some extent counter-intuitive since some noisy networks ($\sigma_{\textrm{val}}>0$) outperform the conventionally trained clean network  ($\sigma_{\textrm{train}}=\sigma_{\textrm{val}}=0$) even the training noise and validation noise are mismatched ($\sigma_{\textrm{train}}\neq \sigma_{\textrm{val}}$). This is observed in vanilla RNNs and the stroke-based MNIST dataset as well.

%\begin{figure}[h]
%	\centering
%	\includegraphics[width=1\linewidth]{lstm_mnist_noisy_training}
%	\caption{Prediction accuracy and error rate of LSTMs with Deep Noise Injection training of $\sigma_{\textrm{train}}=0$ to $1.0$ at a step size of $0.1$. Validated on $\sigma_{\textrm{val}}=0$ to $1.0$ at a step size of $0.1$. }
%	\label{fig:lstm_mnist_noisy_training}
%\end{figure}

\begin{figure}[h]
	\centering
	\includegraphics[width=1\linewidth]{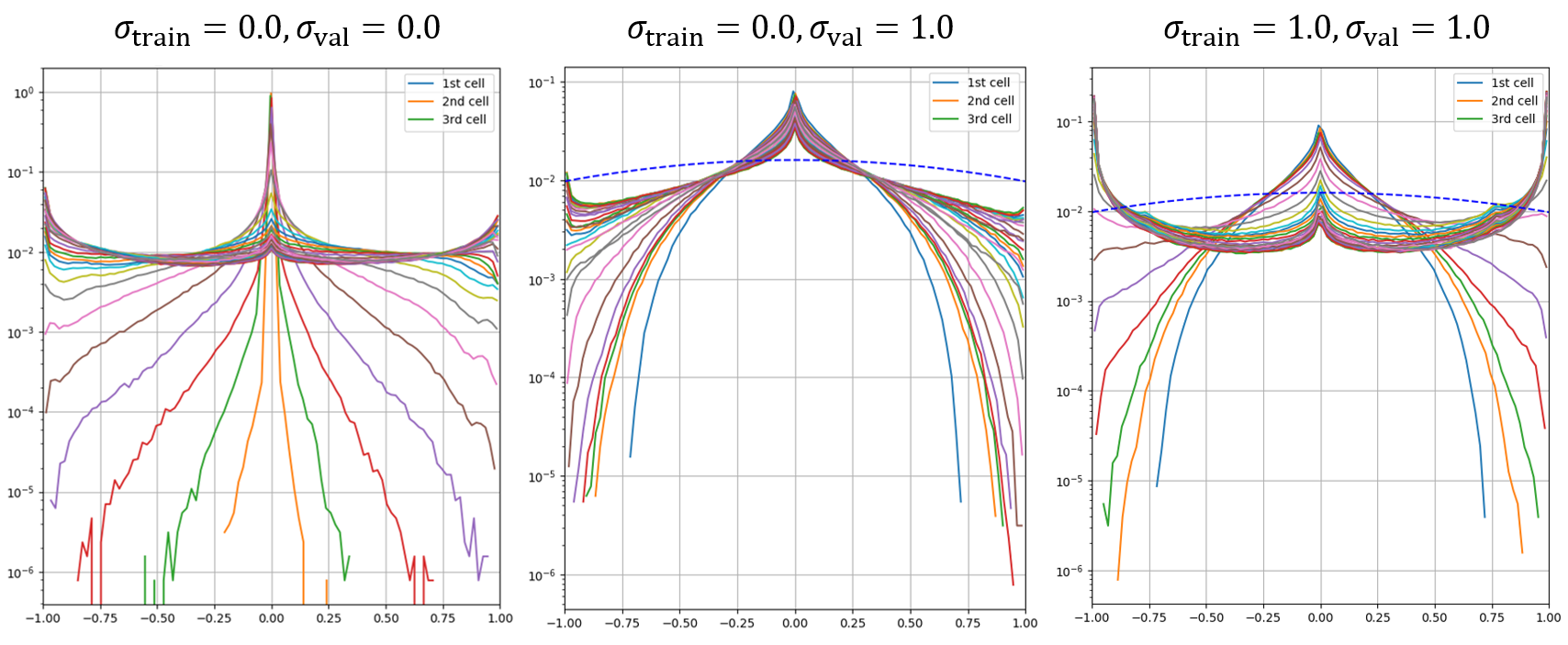}
	\caption{Distributions of the LSTM output states for 28 LSTM cells}
	\label{fig:lstm_state_a}
\end{figure}

\begin{figure}[h]
	\centering
	\includegraphics[width=1\linewidth]{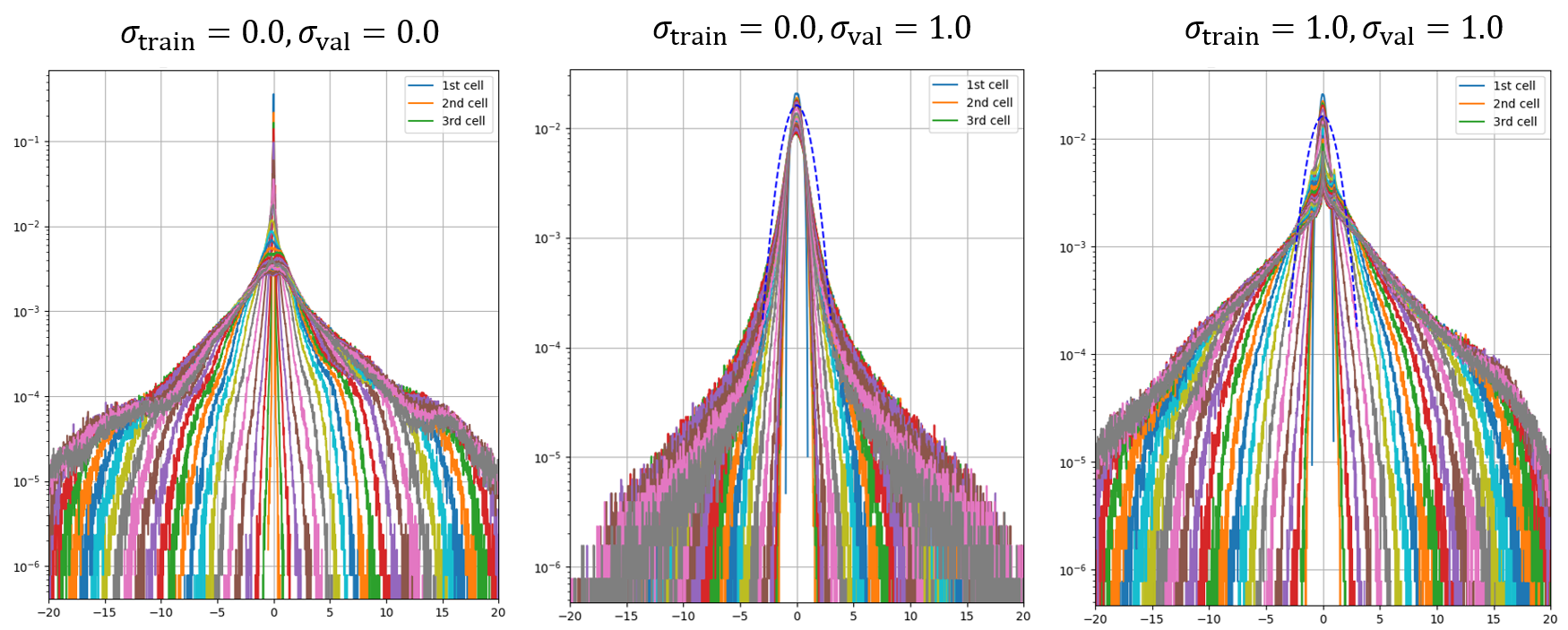}
	\caption{Distributions of the LSTM internal states for 28 LSTM cells}
	\label{fig:lstm_state_c}
\end{figure}

\begin{figure}[htbp]
	\centering
	\includegraphics[width=0.7\linewidth]{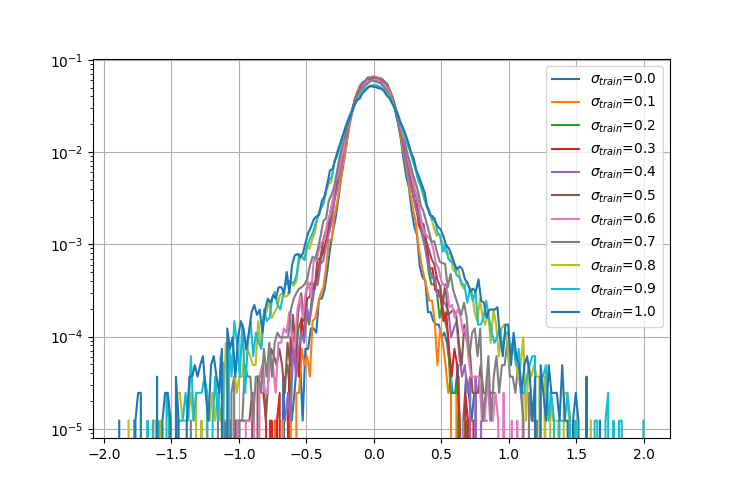}
	\caption{Histogram of weights for conventionally trained LSTM model ($\sigma_{\textrm{train}}=0$) and 10 Deep Noise Injection trained LSTM models ($\sigma_{\textrm{train}}=0.1$ to $1.0$ with a step size of $0.1$).}
	\label{fig:hist_lstm_weights_tr00to10}
\end{figure}

Fig.~\ref{fig:lstm_state_a} and Fig.~\ref{fig:lstm_state_c} show the signal distributions of the output states (lower horizontal arrow connecting two LSTM cells in Fig.~\ref{fig:lstm_cell}) and internal states (upper horizontal arrow connecting two LSTM cells in Fig.~\ref{fig:lstm_cell}) of 28 LSTM cells for three $(\sigma_{\textrm{train}},\sigma_{\textrm{val}})$ pairs. The blue dashed lines show the distributions of the Gaussian noise $z\sim \cN(0,1)$. It is truncated in Fig.~\ref{fig:lstm_state_a} at $[-1,1]$ since the output states are the ouput of a ``tanh'' function and bounded between $\pm 1$ . Only the first three cells have legends due to space limitations in the figures, but we can observe that the signals are more spread in later LSTM cells. The rightmost histogram is visually closer than the middle one to the leftmost histogram, which might explain the huge difference in their classification accuracy ($98\%$ vs $12.5\%$). Fig.~\ref{fig:hist_lstm_weights_tr00to10} shows the distributions of weights for the 10 Deep Noise Injection LSTM models and 1 conventionally trained LSTM model. It can be observed that weights after Deep Noise Injection training are generally spread wider and thus have more power to combat the noise: the expected power (average of the squared values) of all weights increases from $0.013$ ($\sigma_{\textrm{train}}=0$) to $0.037$ ($\sigma_{\textrm{train}}=1.0$).


\subsection{A 50-step LSTM architecture with Input Dim = 2 for MNIST}
\begin{figure}[h]
	\centering
	\includegraphics[width=0.7\linewidth]{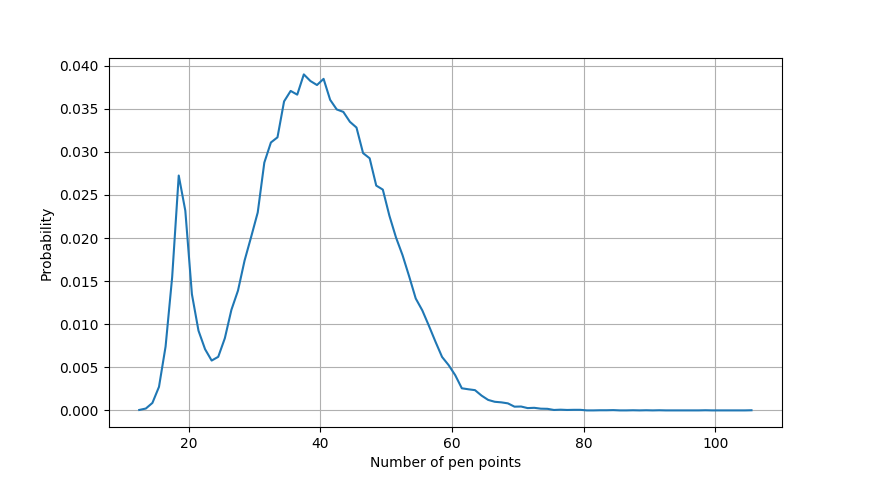}
	\caption{Histogram of the length of stroke sequences (number of pen points) of MNIST training data}
	\label{fig:hist_pen_points}
\end{figure}

In this section we use a sequence of 2-D coordinates to represent the stroke sequence of a handwritten digit image in MNIST. The coordinates are obtained by~\cite{mnist_stroke}. Thus, the input size to a LSTM cell is 2 (compared to 28 in Table~\ref{tab:lstm_mnist_input28}). We use 50 LSTM cells to track the first 50 pen points of the handwritten images. If the total number of pen points (length of the stroke sequences) is less than 50, they are padded with $0$s. The number of LSTM steps being $50$ is empirically chosen to speed up training and obtain reasonable accuracy. Choosing a larger number of LSTM cells (e.g., $70$) results in a low likelihood of convergence at higher $\sigma_{\textrm{train}}$. 
%The size of internal and output states are still 128 and the total number of trainable parameters is $68362$. 
The LSTM architecture is summarized in Table~\ref{tab:lstm_mnist_input2} and the histogram of the length of the stroke sequences is shown in Fig.~\ref{fig:hist_pen_points} where the majority of length resides between $20$ to $60$.
\begin{table}[htbp]
	\center
	\caption{A 50-step LSTM architecture for MNIST}
	\label{tab:lstm_mnist_input2}
	\begin{tabular}{|c|c|}
		\hline No. LSTM cells & 50 \\
		\hline Input shape & (2,1) \\
		\hline LSTM internal state size & 128\\
		\hline LSTM output state size & 128 \\
		\hline Total No. trainable parameters & $68362$\\
		\hline
	\end{tabular} 
\end{table}

Fig.~\ref{fig:lstm_mnist_image_and_stroke_noisy_training}$(b)$ shows the prediction accuracy based on the stroke sequences of MNIST. There are $11\times 11$ $(\sigma_{\textrm{train}},\sigma_{\textrm{val}})$ pairs with $\sigma_{\textrm{train}}=0.0$ to $1.0$ and $\sigma_{\textrm{val}}=0.0$ to $1.0$, both with a step size of $0.1$. Each point is obtained as the average of $40$ independent tests. The top curve corresponds to conventional training without noise injection. It shows the least robustness against noisy computations: validation accuracy drops from $96\%$ to $10.5\%$ as the noise power increases from $0.0$ to $1.0$.
Increasing $\sigma_{\textrm{train}}$ generally results in higher robustness; in particular, when $\sigma_{\textrm{train}}=0.9$ validation accuracy at all noise levels during inference is greater than $92\%$.
It is expected that prediction accuracy based on stroke sequences is not as good as that based on images since padding and truncating the stroke sequences adds unnecessary information and reduces useful information relative to the content of images. Similar to LSTMs in Fig.~\ref{fig:lstm_mnist_image_and_stroke_noisy_training}$(a)$, training with noise injection ($\sigma_{\textrm{train}}>0$) sometimes outperforms conventional training ($\sigma_{\textrm{train}}=0$) even when the validation has no noise in computation ($\sigma_{\textrm{val}}=0$), as can be observed by those points below the red horizontal line.

%\begin{figure}[h]
%	\centering
%	\includegraphics[width=1\linewidth]{lstm_mnist_stroke_noisy_training}
%	\caption{Prediction accuracy and error rate of LSTMs for stroke sequences with Deep Noise Injection training of $\sigma_{\textrm{train}}=0$ to $1.0$ at a step size of $0.1$. Validated on $\sigma_{\textrm{val}}=0$ to $1.0$ at a step size of $0.1$. }
%	\label{fig:lstm_mnist_stroke_noisy_training}
%\end{figure}

\begin{figure}[h]
	\centering
	\includegraphics[width=1\linewidth]{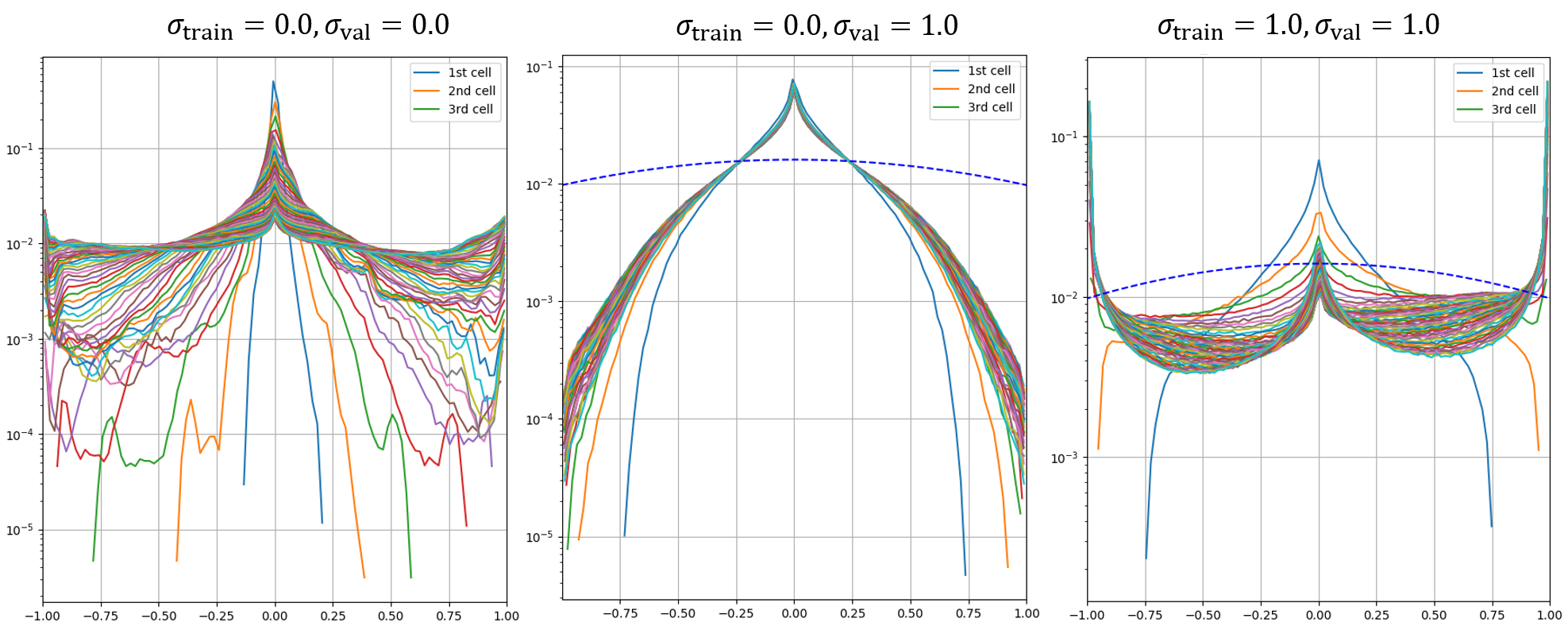}
	\caption{Distributions of the LSTM output states for 50 LSTM cells}
	\label{fig:stroke_lstm_state_a3cases}
\end{figure}

\begin{figure}[h]
	\centering
	\includegraphics[width=1\linewidth]{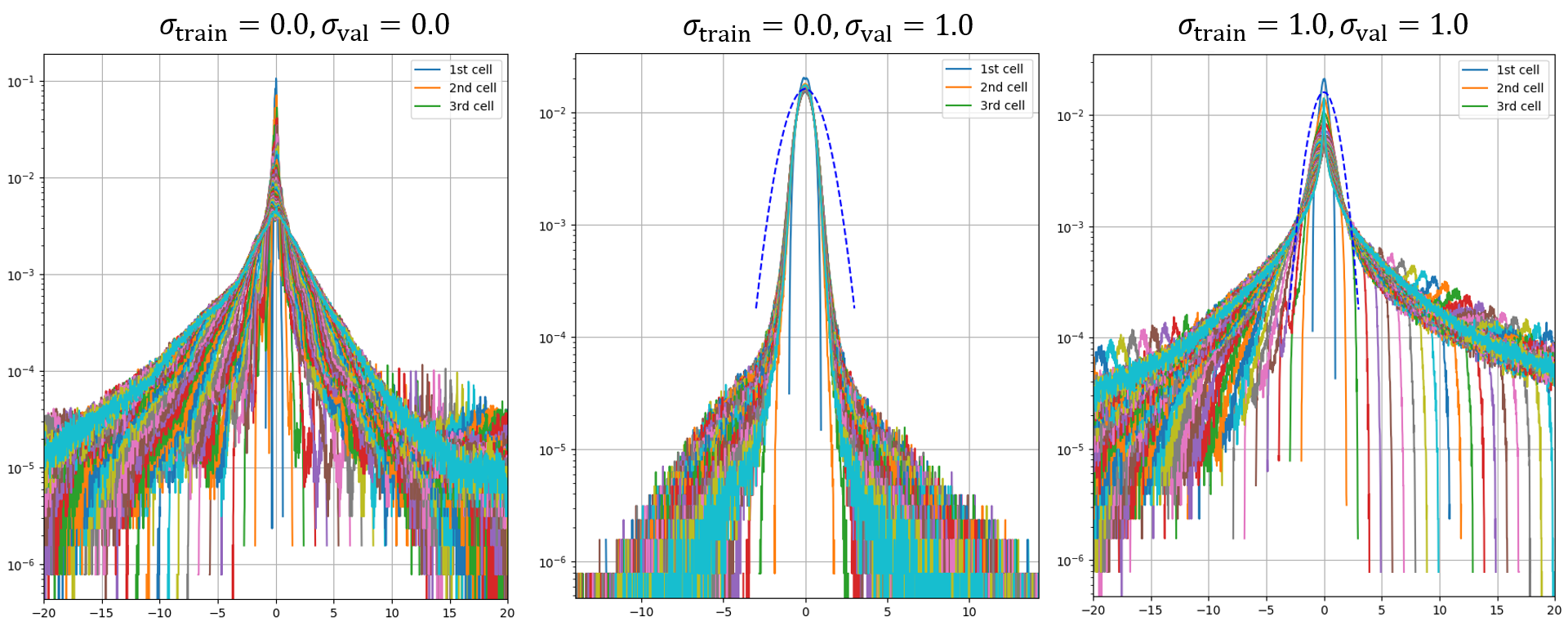}
	\caption{Distributions of the LSTM internal states for 50 LSTM cells}
	\label{fig:stroke_lstm_state_c3cases}
\end{figure}

\begin{figure}[h]
	\centering
	\includegraphics[width=0.7\linewidth]{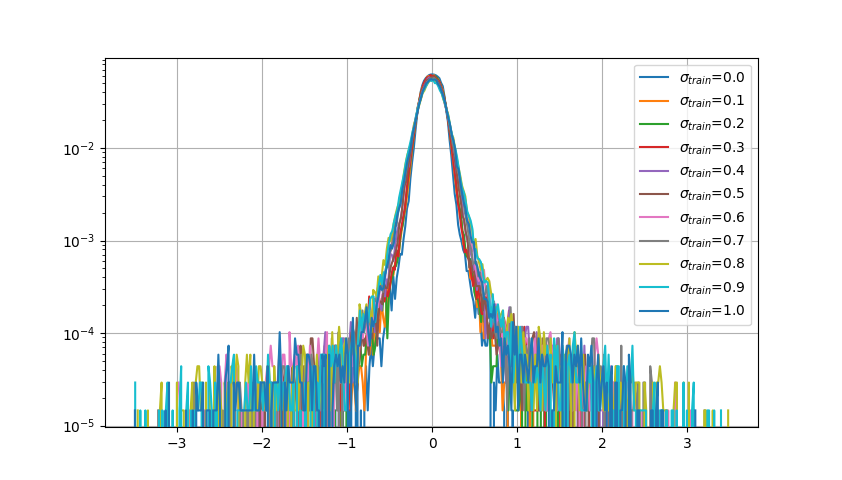}
	\caption{Histogram of weights for conventionally trained LSTM model ($\sigma_{\textrm{train}}=0$) and 10 Deep Noise Injection trained LSTM models ($\sigma_{\textrm{train}}=0.1$ to $1.0$ with a step size of $0.1$) based on stroke sequences of MNIST.}
	\label{fig:hist_stroke_lstm_weights_tr00to10}
\end{figure}

Fig.~\ref{fig:stroke_lstm_state_a3cases} and Fig.~\ref{fig:stroke_lstm_state_c3cases} show the distributions of signals of the output states and internal states of 50 LSTM cells for three $(\sigma_{\textrm{train}},\sigma_{\textrm{val}})$ pairs. The blue dashed lines shows the distributions of the Gaussian noise $z\sim \cN(0,1)$. Only the first three cells have legends due to space limitations in the figures, but we can observe that the signals are more spread in later LSTM cells. Fig.~\ref{fig:hist_stroke_lstm_weights_tr00to10} shows the distributions of weights for the 10 Deep Noise Injection LSTM models and 1 conventionally trained LSTM model. Similar to image-based LSTMs, the weights for Deep Noise Injection training are generally spread wider and thus have more power to combat the noise. The expected power of all weights increases from $0.018$ ($\sigma_{\textrm{train}}=0$) to $0.042$ ($\sigma_{\textrm{train}}=1.0$).

\subsection{A 28-step vanillar RNN architecture with Input Dim = 28 for MNIST}
LSTMs are the most widely used component of recurrent neural networks. As a supplement, we also explore the robustness of vanilla RNNs against noisy computation. The most important observations are the same as the LSTMs, where Deep Noise Injection training largely improves the validation accuracy with the presence of noise. Table~\ref{tab:rnn_mnist} summarizes the RNN architecture used in our experiment.

\begin{table}[htbp]
	\center
	\caption{A 28-step vanilla RNN architecture for MNIST}
	\label{tab:rnn_mnist}
	\begin{tabular}{|c|c|}
		\hline No. RNN cells & 28 \\
		\hline Input shape & (28,1) \\
		\hline No. neurons in one RNN cell (RNN state size) & 128\\
		\hline Total No. trainable parameters & 21386\\
		\hline
	\end{tabular} 
\end{table}

\begin{figure}[h]
	\centering
	\includegraphics[width=0.8\linewidth]{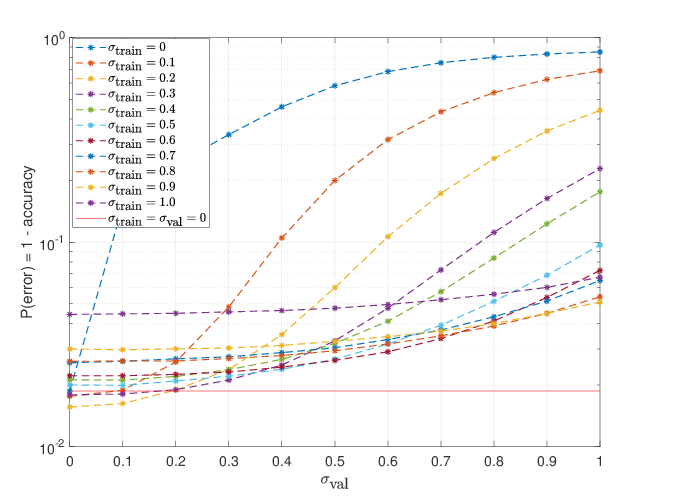}
	\caption{Prediction accuracy and error rate of vanilla RNNs for stroke sequences with Deep Noise Injection training of $\sigma_{\textrm{train}}=0$ to $1.0$ at a step size of $0.1$. Validated on $\sigma_{\textrm{val}}=0$ to $1.0$ at a step size of $0.1$. }
	\label{fig:rnn_mnist_noisy_training}
\end{figure}

Fig.~\ref{fig:rnn_mnist_noisy_training} shows the prediction accuracy of $(\sigma_{\textrm{train}},\sigma_{\textrm{val}})$ pairs with $\sigma_{\textrm{train}}=0.0$ to $1.0$ and $\sigma_{\textrm{val}}=0.0$ to $1.0$, both with a step size of $0.1$. Each point is obtained as the average of $40$ independent tests. The top curve corresponds to conventional training without noise injection. It shows the least robustness against noisy computations: validation accuracy drops from $98\%$ to $15\%$ as the noise power increases from $0.0$ to $1.0$. Increasing $\sigma_{\textrm{train}}$'s generally results in improved robustness; in particular, when $\sigma_{\textrm{train}}=0.9$ validation accuracy at all noise levels during inference is greater than $94\%$. 

%Fig.~\ref{fig:rnn_state_3cases} shows the distributions of signals of the states of 28 vanillar RNN cells for three $(\sigma_{\textrm{train}},\sigma_{\textrm{val}})$ pairs. The blue dashed lines shows the distributions of the Gaussian noise $z\sim \cN(0,1)$. Only the first three cells have legends due to space limitations in the figures, but we can observe that the signals are more spread in later RNN cells. Fig.~\ref{fig:hist_rnn_weights_tr00to10} shows the distributions of weights for the 10 Deep Noise Injection vanilla RNN models and 1 conventional trained vanilla RNN model. The weights for Deep Noise Injection training are generally spread wider and thus have more power to combat the noise. The expected power of all weights increases from $0.014$ ($\sigma_{\textrm{train}}=0$) to $0.06$ ($\sigma_{\textrm{train}}=1.0$).
%
%\begin{figure}[h]
%	\centering
%	\includegraphics[width=1\linewidth]{rnn_state_3cases}
%	\caption{Distributions of the RNN states for 28 RNN cells}
%	\label{fig:rnn_state_3cases}
%\end{figure}
%\begin{figure}[h]
%	\centering
%	\includegraphics[width=1\linewidth]{hist_rnn_weights_tr0_0_to_1_0}
%	\caption{Histogram of weights for conventionally trained RNN model ($\sigma_{\textrm{train}}=0$) and 10 Deep Noise Injection trained RNN models ($\sigma_{\textrm{train}}=0.1$ to $1.0$ with a step size of $0.1$).}
%	\label{fig:hist_rnn_weights_tr00to10}
%\end{figure}
%
%
%\section{Further Discussions}

\section{Conclusions}
In this paper we explored the behavior of RNNs under noisy computations, such as those resulting from analog noise in neuromorphic circuits.
We show that by injecting noise after each matrix-vector multiplication during training, we make a worthy tradeoff between validation accuracy of the noiseless case and high power noise cases. Experiments on the MNIST dataset reveal that with the presence of noise during computation and for all test RNN architectures, including LSTMs and vanilla RNNs, validation accuracy can be improved from ($12.5\%,10.5\%, 15\% $) to over $(98\%, 92\%, 94\%)$, respectively. In addition, Deep Noise injection can improve slightly the overall performance of all three recurrent networks we studied here even when inference is numerically accurate.

\bibliographystyle{IEEEtran}
\bibliography{reference_minghai}

\begin{thebibliography}{10}
\providecommand{\url}[1]{#1}
\csname url@samestyle\endcsname
\providecommand{\newblock}{\relax}
\providecommand{\bibinfo}[2]{#2}
\providecommand{\BIBentrySTDinterwordspacing}{\spaceskip=0pt\relax}
\providecommand{\BIBentryALTinterwordstretchfactor}{4}
\providecommand{\BIBentryALTinterwordspacing}{\spaceskip=\fontdimen2\font plus
\BIBentryALTinterwordstretchfactor\fontdimen3\font minus
  \fontdimen4\font\relax}
\providecommand{\BIBforeignlanguage}[2]{{%
\expandafter\ifx\csname l@#1\endcsname\relax
\typeout{** WARNING: IEEEtran.bst: No hyphenation pattern has been}%
\typeout{** loaded for the language `#1'. Using the pattern for}%
\typeout{** the default language instead.}%
\else
\language=\csname l@#1\endcsname
\fi
#2}}
\providecommand{\BIBdecl}{\relax}
\BIBdecl

\bibitem{Schmidhuber15}
J.~Schmidhuber, ``Deep learning in neural networks: An overview,'' \emph{Neural
  Networks}, vol.~61, pp. 85--117, 2015, published online 2014; based on TR
  arXiv:1404.7828 [cs.NE].

\bibitem{HTF01}
T.~Hastie, R.~Tibshirani, and J.~Friedman, \emph{The Elements of Statistical
  Learning}.\hskip 1em plus 0.5em minus 0.4em\relax Springer, New York, 2001.

\bibitem{GLF09}
A.~Graves, M.~Liwicki, S.~Fernandez, R.~Bertolami, H.~Bunke, and
  J.~Schmidhuber, ``A novel connectionist system for improved unconstrained
  handwriting recognition,'' \emph{IEEE Trans. Pattern Analysis and Machine
  Intelligence}, vol.~31, no.~5, pp. 855 -- 868, May 2009.

\bibitem{HAF14}
H.~Sak, A.~Senior, and F.~Beaufays, ``Long short-term memory recurrent neural
  network architectures for large scale acoustic modeling,'' 2014.

\bibitem{GSKSS16}
K.~Greff, R.~Srivastava, J.~Koutník, B.~Steunebrink, and J.~Schmidhuber,
  ``{LSTM}: A search space odyssey,'' \emph{IEEE Trans. Neural Networks and
  Learning Systems}, pp. 1--11, July 2016.

\bibitem{RB93}
M.~Riedmiller and H.~Braun, ``A direct adaptive method for faster
  backpropagation learning: the {RPROP} algorithm,'' in \emph{IEEE Int. Conf.
  Neural Networks}, San Francisco, CA, USA, March 2009, pp. 586--591.

\bibitem{Nielsen89}
R.~Hecht-Nielsen, ``Theory of the backpropagation neural network,'' in
  \emph{Int. Joint Conf. on Neural Networks (IJCNN)}, Washington, DC, USA,
  1989, pp. 593--605.

\bibitem{ZADFBG10}
W.~Zhao, G.~Agnus, V.~Derycke, A.~Filoramo, J.~Bourgoin, and C.~Gamrat,
  ``Nanotube devices based crossbar architecture: toward neuromorphic
  computing,'' \emph{Nanotechnology}, vol.~21, no.~17, April 2010.

\bibitem{FWI18}
B.~Feinberg, S.~Wang, and E.~Ipekn, ``Making memristive neural network
  accelerators reliable,'' pp. 52--65, 02 2018.

\bibitem{JRRR17}
S.~Jain, A.~Ranjan, K.~Roy, and A.~Raghunathan, ``Computing in memory with
  spin-transfer torque magnetic ram,'' \emph{IEEE Transactions on Very Large
  Scale Integration (VLSI) Systems}, vol.~23, no.~3, pp. 470--483, December
  2017.

\bibitem{CW17}
N.~Carlini and D.~Wagner, ``Towards evaluating the robustness of neural
  networks,'' in \emph{IEEE Symp. on Security and Privacy (SP)}, San Jose, CA,
  USA, 2017, pp. 39--57.

\bibitem{ZSLG16}
S.~Zheng, Y.~Song, T.~Leung, and I.~Goodfellow, ``Improving the robustness of
  deep neural networks via stability training,'' in \emph{IEEE Conf. on
  Computer Vision and Pattern Recognition (CVPR)}, Las Vegas, NV, USA, 2016,
  pp. 4480 -- 4488.

\bibitem{YLZetal2015}
\BIBentryALTinterwordspacing
S.~Yin, C.~Liu, Z.~Zhang, Y.~Lin, D.~Wang, J.~Tejedor, T.~F. Zheng, and Y.~Li,
  ``Noisy training for deep neural networks in speech recognition,''
  \emph{EURASIP Journal on Audio, Speech, and Music Processing}, vol. 2015,
  no.~1, p.~2, January 2015. [Online]. Available:
  \url{https://doi.org/10.1186/s13636-014-0047-0}
\BIBentrySTDinterwordspacing

\bibitem{Andrzej14}
A.~Rusiecki, M.~Kordos, T.~Kami{\'{n}}ski, and K.~Gre{\'{n}}, ``Training neural
  networks on noisy data,'' in \emph{Artificial Intelligence and Soft
  Computing}.\hskip 1em plus 0.5em minus 0.4em\relax Cham: Springer
  International Publishing, 2014, pp. 131--142.

\bibitem{HSMDS00}
R.~Hahnloser, R.~Sarpeshkar, M.~Mahowald, R.~J. Douglas, and H.~S. Seung,
  ``Digital selection and analogue amplification coexist in a cortex-inspired
  silicon circuit,'' \emph{Nature}, no. 405, p. 947–951, 2000.

\bibitem{MHN13}
A.~L. Maas, A.~Y. Hannun, and A.~Y. Ng., ``Rectifier nonlinearities improve
  neural network acoustic models,'' 2013.

\bibitem{mnist_stroke}
E.~D. de~Jong, ``{MNIST digits stroke sequence data},''
  \url{https://github.com/edwin-de-jong/mnist-digits-stroke-sequence-data.wiki.git},
  2014, [Online; accessed 07-March-2018].

\end{thebibliography}

\end{document}